\documentclass[conference]{IEEEtran}
\IEEEoverridecommandlockouts
\usepackage{cite}
\usepackage{amsmath,amssymb,amsfonts}
\usepackage{algorithmic}
\usepackage{graphicx}
\usepackage{textcomp}
\usepackage{xcolor}
\usepackage{multirow}
\usepackage[noabbrev]{cleveref}
\usepackage{pbox}
\usepackage{adjustbox}
\usepackage{array}
\usepackage{rotating}
\usepackage[flushleft]{threeparttable} 

\usepackage{tikz}
\usepackage{amssymb}
\usepackage{wasysym}

\usepackage{textcomp}
\usepackage{fancyhdr}

\usepackage[utf8]{inputenc}

\usepackage{tikz}
\usepackage{textcomp}

\newcommand\copyrighttext{%
  \footnotesize \textcopyright 2020 IEEE. Personal use of this material is permitted.
  Permission from IEEE must be obtained for all other uses, in any current or future
  media, including reprinting/republishing this material for advertising or promotional
  purposes, creating new collective works, for resale or redistribution to servers or
  lists, or reuse of any copyrighted component of this work in other works.}
\newcommand\copyrightnotice{%
\begin{tikzpicture}[remember picture,overlay]
\node[anchor=south,yshift=10pt] at (current page.south) {\fbox{\parbox{\dimexpr\textwidth-\fboxsep-\fboxrule\relax}{\copyrighttext}}};
\end{tikzpicture}%
}

\makeatletter

\@ifundefined{showcaptionsetup}{}{
 \PassOptionsToPackage{caption=false}{subfig}}
\usepackage{subfig}
\makeatother

\usepackage{eso-pic}
\newcommand\AtPageUpperMyright[1]{\AtPageUpperLeft{
 \put(\LenToUnit{0.5\paperwidth},\LenToUnit{-1cm}){
     \parbox{0.5\textwidth}{\raggedleft\fontsize{9}{11}\selectfont #1}}
 }}
\newcommand{\conf}[1]{
\AddToShipoutPictureBG*{
\AtPageUpperMyright{#1}
}
}

\conf{Preprint} 

\def\BibTeX{{\rm B\kern-.05em{\sc i\kern-.025em b}\kern-.08em
    T\kern-.1667em\lower.7ex\hbox{E}\kern-.125emX}}
\begin{document}

\title{How to Learn from Others: Transfer Machine Learning with Additive Regression Models to Improve Sales Forecasting
}

\author{
\IEEEauthorblockN{Robin Hirt}
\IEEEauthorblockA{
\textit{Karlsruhe Institute of Technology (KIT)}\\
Karlsruhe, Germany \\
robin.hirt@kit.edu}
\and
\IEEEauthorblockN{Niklas Kühl}
\IEEEauthorblockA{
\textit{Karlsruhe Institute of Technology (KIT)}\\
Karlsruhe, Germany \\
niklas.kuehl@kit.edu}
\and
\IEEEauthorblockN{Yusuf Peker}
\IEEEauthorblockA{
\textit{Karlsruhe Institute of Technology (KIT)}\\
Karlsruhe, Germany \\
office@ksri.kit.edu}
\and
\IEEEauthorblockN{Gerhard Satzger}
\IEEEauthorblockA{
\centerline{\textit{Karlsruhe Institute of Technology (KIT)}}\\
Karlsruhe, Germany \\
gerhard.satzger@kit.edu}
}

\maketitle
\copyrightnotice

\begin{abstract}
In a variety of business situations, the introduction or improvement of machine learning approaches is impaired as these cannot draw on existing analytical models. However, in many cases similar problems may have already been solved elsewhere—but the accumulated analytical knowledge cannot be tapped to solve a new problem, e.g., because of privacy barriers. 

For the particular purpose of sales forecasting for similar entities, we propose a transfer machine learning approach based on additive regression models that lets new entities benefit from models of existing entities.  We evaluate the approach on a rich, multi-year dataset of multiple restaurant branches. We differentiate the options to simply transfer models from one branch to another (“zero shot”) or to transfer and adapt them. We analyze feasibility and performance against several forecasting benchmarks. The results show the potential of the approach to exploit the collectively available analytical knowledge.

Thus, we contribute an approach that is generalizable beyond sales forecasting and the specific use case in particular. In addition, we demonstrate its feasibility for a typical use case as well as the potential for improving forecasting quality. These results should inform academia, as they help to leverage knowledge across various entities, and have immediate practical application in industry.
\end{abstract}


\section{Introduction}

Within the last decade, machine learning has gained increased popularity in solving business-related problems. Various studies show impressive capabilities of machine learning in predictive tasks, e.g., to determine whether customers will buy products \cite{Martens2016a} or how leaders in online communities \cite{Johnson2015a} can be identified. However, the creation and sharing of machine learning models in research and practice still holds many challenges \cite{HirtKuehl2018_1000086441}. When it comes to the access to machine learning models, two issues are prevalent: First, for new problems, elementary training data may only become available over time, while predictions are instantly needed (lack of initial data) \cite{Hashem2014}. Second, as privacy and IP preservation often play an important role \cite{Jensen2013}, valuable data available in other places often cannot be tapped in order to build powerful models (lack of data exchange). 

We regard the concept of transfer machine learning as a promising solution to address both challenges, as it allows to transfer analytical models without exchanging raw data. On the basis of these models, predictions can be made---even when there is a lack of training data. Additionally, its application could allow for more efficiency across different entities in a system (e.g., business units or plants within a company), as the same problem does not need to be solved many times: A once trained model can be re-applied multiple times for similar problems at each entity. While the purely technical capabilities of transfer machine learning have been covered in the field of computer science \cite{Pan2010}, effective and efficient applications to business problems are rare. However, first studies \cite{Hopf2017a} show impressive results, and transfer machine learning could be a powerful method in the toolbox of data scientists. Furthermore, a successful transfer could enable novel “machine learning model markets”, that would support the exchange of generalizable knowledge in the form of encapsulated, “ready-to-deploy” models. Additionally, we observe a lack of studies on transfer learning based on "shallow learning" techniques, such as regression-based algorithms. In contrast to deep learning algorithms, shallow learning techniques pose advantages, such as a higher level of explainability or fewer required training samples \cite{lin2018data}.

Time series forecasting is one of the major application areas of machine learning \cite{Zhang2003}, popular, among others for the prediction of sales in various industries \cite{Schneider2016a, Wacker2002}. For the particular purpose of sales forecasting for similar entities, we propose a transfer machine learning approach that lets new entities benefit from models of existing entities.  We design a novel approach on the basis of transfer machine learning for sales forecasting, which we apply to a unique and extremely rich dataset from different restaurant chains and their associated individual outlets as entities (“restaurants”). We build and analyze models at two different levels of adaption: a “zero shot” solution as introduced by Hopf et al. (2017) \cite{Hopf2017a}, which simply transfers models from other entities, as well as a newly contributed solution which further adapts transferred models to the data of the entity in question. The approaches are able to deliver better results than appropriate benchmarks, and, thus, can be a viable means to address the two challenges of lack of initial data as well as lack of data exchange. Thus, we contribute an approach that is well generalizable beyond sales forecasting, and demonstrate its feasibility for a typical use case as well as the potential for improving forecasting quality. A particularly attractive implication would be to allow for sharing of models between different legal entities to enable better, system-wide analyses—without risking the exposure of sensitive data.

The paper at hand is structured as follow: First, we present our research questions. In the third chapter, we cover related work. Then, we introduce the use case, present the research design and elaborate on performance metrics. We then evaluate the feasibility and performance of “isolated” sales forecasting for individual restaurants—without transferring models. These results then serve as a baseline for the transferred models in the fourth section. There, we transfer models between restaurants with different degrees of transfer machine learning—and report on the performances. Finally, we summarize the results, discuss their generalization, recognize limitations, and show future research prospects.

\section{Research Questions}

As a first aspect of this work, we aim to design, implement and evaluate a method for sales forecasting in the area of restaurant food chains. Hereby, we focus on each branch separately and do not exchange any data or models across them. To evaluate the quality of our results, we also implement trivial models as a baseline. Thus, we state the first research question (RQ):

\begin{center}
   \textit{RQ 1: How well can we forecast sales for single stores based on historic data?} 
\end{center}

If we can achieve superior statistical results in comparison to baseline models for one branch, the question arises, if and how we are able to learn from data of other branches to gain analytical knowledge and, thus, improve the forecasting performance. Therefore, we evaluate if it is possible to transfer an analytical model from one branch (source) to another branch (target). A transfer could be beneficial in many ways. In cases, where no historic data is available to build a prediction model, the application of a previously built model would make a prediction possible. This type of prediction problem is called “zero shot” \cite{Fu2015} and is especially relevant for branches without historic data available, e.g., new branches or branches in planning. Thus, we ask the following research question:

\begin{center}
    \textit{RQ 2: Can we transfer models from one branch (source) to a different one (target) to forecast sales—without having any historic data of the target branch available (“zero shot”)? }  
\end{center}

For branches, where little historic sales streams are available, it could be beneficial to profit from a larger historic data set that originates from another branch to learn from patterns that both branches have in common. Thus, we aim to examine the performance of models that are pre-trained on data of one store (source) and then transferred and further adapted on data of another store (target):

\begin{center}
    \textit{RQ 3: How well can we forecast sales with adapted transferred models?}
\end{center}


\section{Related Work \& Contribution to Theory}

Related work can be structured along two major research areas. We first give an outline on time series prediction with the particular application to sales forecasting and describe common techniques in that area. Then, transfer learning and applications in different domains are described. Finally, we demarcate this work from related literature and state its contribution.

\subsection{Time Series Prediction \& Sales Forecasting}

Time series prediction is of importance in many fields, such as water quality prediction \cite{Omer2010} or customer demand forecasting \cite{Lasek2016a}. A time series consists of many consecutive observations that can each be linked to a certain time stamp \cite{Brockwell1991}. To forecast time series, historic data is used to build models, describing patterns that appear in the time series to predict future values \cite{Zhang2003}. There are different approaches towards developing a forecasting model, such as applying an autoregressive integrated moving average (ARIMA) model \cite{Arunraj2015}, machine learning models based on support vector machines \cite{Sapankevych2009}, artificial neural networks (ANN) \cite{Kaastra1996}, or hybrid approaches between machine learning techniques and ARIMA \cite{Arunraj2015, Gao2017, Omer2010, Zhang2003}. Especially for predicting sales in retail or the hospitability industry, time series forecasting—as sales forecasting—can yield various advantages and depicts one important pillar for success \cite{Lasek2016,Arunraj2015,Celebi2017}.


\subsection{Transfer Machine Learning}

As forecasting models in general “learn” from historic data points to distinguish and predict certain instances in the future, our goal is to transfer those learnings from one problem to another. The domain of transfer machine learning has a wide range of applications in real world domains, such as natural language processing or computer vision \cite{Lu2015}. For example, in the area of object recognition in computer vision, a pre-trained model on one domain can improve prediction accuracy—especially, when there is little training data in the target \cite{Shin2016}. In general, transfer learning “is used to improve a learner from one domain by transferring information from a related domain” \cite{Weiss2016}. Similar to Pan and Yang (2010) \cite{Pan2010}, we further specify in this work, that a learner has an originating source domain and a source task and may benefit from a target domain and a target task. 

Pan and Yang (2010) \cite{Pan2010} distinct between various learning settings, depending on if and how the source and target domain and task of the considered approach differ: inductive, unsupervised and transductive transfer machine learning. In our case, we want to transfer a model from one sales forecasting task to another (same task) where the source and target domain of our setting are different, but related (inductive transfer learning). 
Furthermore, Pan and Yang (2010) \cite{Pan2010} distinguish between three main questions that research aims to answer in the area of transfer machine learning: what, how and when to transfer. 
Correspondingly, in this work, we aim to find out, when a transferred forecasting model performs well. Additionally, we evaluate certain levels of adaption to the target task and domain and, thus, try to answer the question of how to transfer a forecasting models in order to get better performance.

\subsection{Transfer Machine Learning for Sales Forecasting \& Positioning of this Work}

Transfer Machine learning is already applied in different domains to reuse models within a new domain that are originating from another one. Hopf et al. (2017) \cite{Hopf2017a} aim to classify household characteristics—such as number of residents, space or water heating type—by analyzing corporate data with open government data using a decision tree. Additionally, they describe a transfer of their classification model from one country to another. In some cases, the transferred model performs well in the target domain compared to a model solely trained on the target domain’s data. However, Hopf et al. (2017) \cite{Hopf2017a} are describing a classification problem and evaluate a transfer without any adaption (“zero shot”) on two separate sets of data. 

In the area of oil price forecasting, Xiao et al. (2012, 2017) \cite{Xiao2012,Xiao2017} describe an approach on predicting prices on two different types of oil. They try to analyze the similarity of those two and then determine, if it makes sense to additionally use a subset of one data set of an oil price to train a model to predict the other one. However, their source and target data streams are highly interdependent, and they only consider two sets of oil prices. 

In contrast to related work in the area of sales forecasting, we analyze a unique data set composed of sales data originating from different restaurant chains in different cities. In contrast to prior work in transfer learning, such as Hopf et al. (2017) \cite{Hopf2017a}, we consider a time series forecasting rather than a classification problem. Additionally, in contrast to a transfer between only two sets — e.g., Xiao et al. (2012, 2017) \cite{Xiao2012,Xiao2017} — we are able to evaluate the performance of transfer learning on different levels of relationship between data sets as we aim to transfer time series forecasting models between branches of one chain, and even across restaurant chains. Furthermore, there is a lack of studies showing transfer learning based on shallow learning techniques, such as regression-based algorithms. Whereas deep learning based algorithms represent black box methods with little to no means of explainability, techniques, such as a regression yield potential in their simplicity and interpretability \cite{lin2018data}. To further examine the suitability of transfer learning in certain cases and explain our results, we cross-analyze each branch towards their suitability for transfer learning. Furthermore, we consider different levels of adaption of a transferred model to a target branch—which has not been done previously.

Thus, our contribution is threefold:

\begin{itemize}
\item We show the feasibility of a sales forecasting approach on a unique data set composed of two restaurant chains and six branches from 2012 to 2017
\item We show the feasibility of a novel approach leveraging transfer machine learning in a sales forecasting case using regression-based algorithms
\item We show the suitability of transfer learning in general and extend its capabilities by using adaptive learning to reach superior performances using regression-based algorithms
\end{itemize}

\section{Use Case: Sales Forecasting for Restaurant Chains}

In this work, we aim to realize sales forecasting on a daily basis for two restaurant chains with different branches. The restaurant chains are serving different types of food and, thus, their sales are following different patterns. By knowing the sales for each branch per day in the next week, month, or even year, several advantages can be leveraged. Based on the revenue and demand, staff schedules can be optimized towards cost savings and a better experience for customers can be delivered. Additionally, the procurement of supplies can be optimized, as spoiled food is a main cost-driver for restaurants. Thus, the management of restaurant chains has a major interest to forecast sales for their branches. In this chapter, we are considering sales forecasting for each branch separately to first show the feasibility and performance of forecasting on our data set and then examine the possibilities of transfer learning in our case.


\subsection{Data set}

Our data set consists of all sales data, including item text, price and amount for two restaurant chains with three branches each. The branches are located in different cities and sales data dates back 48 to 72 months until the end of 2017. Table \ref{table2} gives an overview of chains, branches, start and end date as well as duration of the corresponding data set. We can see that the smallest set of data consists of four years of data (branch 3). For other branches we obtain five (branch 2, 4-6) or six years (branch 1) of sales data. For reasons of confidentiality and comparability, we only disclose normalized data in this study.

\begin{table}[htbp]
\caption{Overview of branches and available sales data.}\label{tab1}
\begin{tabular}{|r|l|l|l|l|l|}
\hline
Branch \# & \textbf{Chain} & \textbf{City} & \textbf{Start} & \textbf{End} & \textbf{Duration} \\
\hline
1                                      & $\alpha$              & A             & 01/01/2012     & 12/31/2017   & 72 \\
2                                      & $\alpha$              & A             & 01/01/2013     & 12/31/2017   & 60 \\
3                                      & $\alpha$              & B             & 01/01/2014     & 12/31/2017   & 48 \\
4                                      & $\beta$              & A             & 01/01/2013     & 12/31/2017   & 60 \\
5                                      & $\beta$              & C             & 01/01/2013     & 12/31/2017   & 60 \\
6                                      & $\beta$              & D             & 01/01/2013     & 12/31/2017   & 60  \\            
\hline
\end{tabular}
\label{table2}
\end{table}

\begin{figure}[htbp]

    \centering
    Branch 1, chain $\alpha$
	\includegraphics[width=0.8\linewidth]{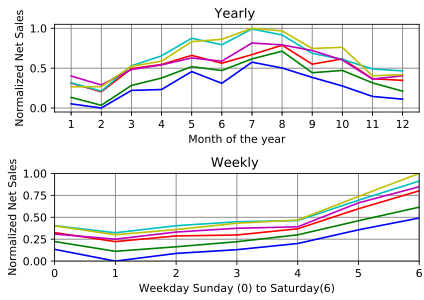}
	\newline
	Branch 4, chain $\beta$
	\includegraphics[width=0.8\linewidth]{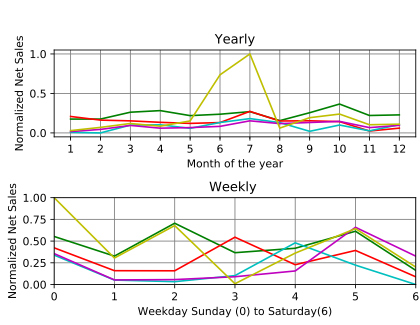}
	\caption{Weekly and yearly seasonality of branch 1, chain $\alpha$ (top) and branch 4, chain $\beta$ (bottom)}
	\label{fig:1}       
\end{figure}

We aggregate daily net sales for each branch and perform an exploratory descriptive analysis to get a better understanding of the structure and seasonality in the sales data time series. Hereby, we observe, that there is a seasonality per week and month. We further analyze the data to better understand patterns that emerge in the time series. In Figure \ref{fig:1}, we depict the weekly and monthly normalized net sales seasonality for two branches 1 and 4, belonging to different chains. For the weekly seasonality, it is noticeable that for branch 1, sales are higher at Friday and Saturday, while on Sunday, sales are low. In case of branch 4, Fridays and Sundays perform very well, but there is a dip on Saturdays. 

Monthly sales on the other hand show similar effects for summer months, as there might be a higher demand for dining in restaurants. One also can see that, while branch 1 has a stable monthly seasonality over historic data, branch 4 only has a weak similarity in monthly net sales. The goal in our approach is to exploit the seasonality in the data to improve our forecasts.



To answer our research questions, we consider four different scenarios (see Table \ref{tab:overview}) to compare retrieved results. The first two scenarios 1a and 1b are addressing RQ 1, scenario 2 is addressing RQ 2 and scenario 3 is addressing RQ 3. In this chapter we aim to answer RQ 1 and, thus, focus on scenario 1a and 1b, where we are not applying transfer learning and solely realize sales forecasting for each branch separately. 

Our data set is composed of sales data from different branches. However, as we do not have the same sales data period for every branch, in scenario 1a, we consider only one year of training (2016) for a branch and one year of testing (2017). The performances of models in this scenario can then be easily compared, as training and test period for each branch are the same. Additionally, we want to show that with more training data models will improve performance. Thus, to fully utilize our data set, in scenario 1b we train our forecasting models on all historic data until the end of 2016 and test on sales data for the year 2017. In scenario 1b, the training period varies between branches.

\begin{table}[htbp]
\caption{Overview of scenarios to evaluate and compare results}
\begin{tabular}{|r|l|l|l|l|l|}
\hline
Scenario & \textbf{\begin{tabular}[c]{@{}l@{}}Training period\\ (source)\end{tabular}} & \textbf{\begin{tabular}[c]{@{}l@{}}Adaption period\\ (target)\end{tabular}} & \textbf{Transfer learning}   \\
\hline
1a  & 2016  & -    & No \\
1b  & Until 2016    & -     & No  \\
2   & Until 2016    & -        & Zero shot    \\
3   & Until 2015    &   2016       & With adaption  \\            
\hline
\end{tabular}
\label{tab:overview}
\end{table}

In the following, we state the metrics for performance evaluation, the considered baselines to compare our results, describe our approach for realizing a sales forecasting model for every restaurant branch separately and, lastly, show the results.

\subsection{Metrics \& Baselines}

In order to evaluate the feasibility of our approach, we choose two common metrics for time series forecast evaluation. First, the Root Mean Square Error (RMSE) which represents the root of the aggregated squared errors of time series forecast as a means to optimize our models. Second, the Mean Average Percentage Error (MAPE) as a means to compare the forecast error among different data sets \cite{Omer2010,Xiao2017}.



For simplicity reasons, in this study, we only show MAPE results. If not stated differently, all results are averaged over 12 months of the forecast period.

To make our results comparable, we also consider a baseline method: the seasonal naïve method \cite{Hyndman2014}, also known as the ‘persistence algorithm’. It forecasts the future by taking the value from the previous year, e.g., the predicted value of February the 3rd in 2017 would be the actual known from February 3rd of 2016. The seasonal naïve method is in fact often used as a baseline for time series approaches \cite{Arunraj2015}.

\section{Sales Forecasting with Isolated Machine Learning}

\subsection{Method}


Our approach consists of different steps and starts with a data cleaning. We pre-process the respective data set and remove noisy data points, such as transactions with wrong time stamps ($<$ 1\%). We also remove tips, as they are no indicator for future sales and could confuse a predictive model. We also define that transactions after midnight until closing are accounted for the previous business day. Additionally, minor fixes are made, such as removing negative sales days that are caused by retrospective corrections of transactions on days without opening hours. 


After that, the data is divided into training and test set (data split). Depending on the scenario, we are considering a training set ranging from 01-01-2016 to 12-31-2016 and a test set ranging from 01-01-2017 to 12-31-2017. Then, we apply a logarithmic function to the base 2, to stabilize the variance of net sales that can appear over time \cite{Box2015}.

Afterwards, the training set composed of logarithmic time series is used to build a forecasting model. For that purpose, many possible algorithms could be applied, such as an ARIMA model \cite{Box2015}, a seasonal ARIMA model \cite{Arunraj2015}, a long short-term neural network model \cite{Chen2015}, an additive regression model \cite{Friedman1981}, or others. The goal of this study is to first show the feasibility of sales forecasting for branches of restaurant chains and then transfer predictive models across those branches. Although we want to obtain good results, the focus on this study lays on evaluating transfer learning to generalize forecasting models. Thus, we are not aiming towards finding the best-possible model, but a well-performing one. For that reason, we pre-tested the approaches mentioned above. 

As a result, for our main model, we use an additive regression model based on three components: a linear trend with changepoints with a constant growth rate. In comparison to black box methods like a neural network, additive regression models allow to visualize the learned components and, thus, interpret them. For implementing the additive regression model, we use the open-source Prophet library \cite{Taylor2017}, which is capable of modelling multi-seasonality combined with a corresponding growth. The simplified linear growth function $g(t)$ can be expressed with $a(t)$ as the adjustments at a time t, $\delta$ as rate adjustments, m as an offset parameter and $\gamma$ as adjustment at a changepoint by the following equation.

\begin{center}
    (3) $g(t) = (k+a(t)^T \delta)t + (m+a(t)^T \gamma$
\end{center}

The growth rate k itself can also change over time, due to a change of demand for certain items or types of food and beverages.  Thus, Prophet allows to detect and set changepoints for the trend data in a time series, that isn't handled by the seasonality calculations. That way, changes in the trajectory of series that might appear can be included by the growth model. However, the further out in the future the model predicts sales, the higher the uncertainty. In our sales data, due to the behavior of customers in different seasons, in different weather or other conditions, a multi-period seasonality appears. This means, that on different levels (e.g., weekly or monthly) patterns in sales data emerge. This multi-seasonality is modelled through a Fourier series \cite{Harvey1993} as described by Taylor and Letham (2017) \cite{Taylor2017}.

After model training, we use the trained model to forecast future sales for the test period. As a result, we get a logarithmic forecast that is transformed by an exponential function. The forecast can now be used in combination with the test set to calculate the performance of the trained model.

\subsection{Results: Isolated Machine Learning}

Our goal is to find out how well we can predict sales for restaurant branches per day based on historic data and, thus, we evaluate two scenarios: one, where we train and test models using sales data from a limited, but comparable period time across branches (1a), and another one, where we use all available data for each branch to fully utilize the data set (1b). Furthermore, we compare our approach with a baseline model. 

\begin{figure}[htbp]
    \centering

    Scenario 1a
	\includegraphics[width=0.8\linewidth]{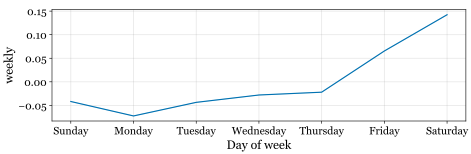}
	\newline 
	Scenario 1b
	\includegraphics[width=0.8\linewidth]{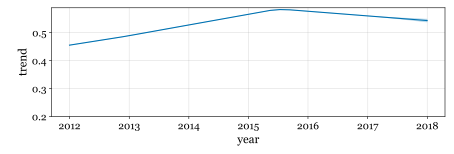}
	\caption{Weekly components of the model for branch 1, chain $\alpha$ in scenario 1a (top) vs. scenario 1b (bottom)}
	\label{fig:3}       
\end{figure}

In addition, we exemplarily show the learned weekly components for branch 1, chain $\alpha$ in both scenario 1a and scenario 1b in Figure 3. In scenario 1a, models are only trained on one year (2016) of historic data. We can observe that the seasonality of the sales series is learned by the additive model. The weekly seasonality, mirror the aggregated data in Figure \ref{fig:3}. 


In order to receive stable results, we test the trained models on a period of one year (2017) to fully understand the performance for a complete year. In contrary, to obtain stable results on a 12-month prediction, the uncertainty of a prediction increases and, thus, the performance decreases. We also test predictions on a period of one and six months after the training period. Overall, we observe, that the further we predict into the future, performance slightly decreases but the overall stability of results increases. The averages of the MAPE in scenario 1a decrease from one month (12.62\%), to six months (21.82\%) and to 12 months (34.12\%). As we are interested in a stable forecasting model and want to minimize random effects, we focus on a 12-month testing period. Thus, unless stated otherwise, in the remainder of this paper all results are referring to a test period of one year (2017). We compare a $performance_1$ with a $performance_2$ by calculating the percentage change as $\frac{performance_1}{performance_2}-1$.

\begin{table}[htbp]
\centering
\scriptsize
\caption{MAPE of baseline method and our method in scenario 1a \& 1b in \%. Comparisons in percentage points.}
\begin{tabular}{r|llllll}
\textbf{Method}                                                        & \textbf{\begin{tabular}[c]{@{}l@{}}Ch. $\alpha$\\ Br. 1\end{tabular}} & \textbf{\begin{tabular}[c]{@{}l@{}}Ch. $\alpha$\\ Br. 2\end{tabular}} & \textbf{\begin{tabular}[c]{@{}l@{}}Ch. $\alpha$\\ Br. 3\end{tabular}} & \textbf{\begin{tabular}[c]{@{}l@{}}Ch. $\beta$\\ Br. 4\end{tabular}} & \textbf{\begin{tabular}[c]{@{}l@{}}Ch. $\beta$\\ Br. 5\end{tabular}} & \textbf{\begin{tabular}[c]{@{}l@{}}Ch. $\beta$\\ Br. 6\end{tabular}} \\ \hline
\textbf{Baseline}                                                      & 21.27                                                              & 26.68                                                              & 42.05                                                              & 26.79                                                              & 32.20                                                              & 29.78                                                              \\
\textbf{1a}                                                            & 26.74                                                              & 26.17                                                              & 33.51                                                              & 36.61                                                              & 29.35                                                              & 52.42                                                              \\
\begin{tabular}[c]{@{}r@{}}Scenario 1a\\ vs. baseline\end{tabular}     & 25.71                                                              & -1.91                                                              & -20.30                                                             & 36.65                                                              & -8.8                                                               & 76.02                                                              \\
\textbf{1b}                                                            & 9.63                                                               & 24.71                                                              & 14.39                                                              & 17.59                                                              & 31.49                                                              & 24.65                                                              \\
\begin{tabular}[c]{@{}r@{}}Scenario 1b\\ vs. baseline\end{tabular}     & -54.72                                                             & -7.38                                                              & -65.77                                                             & -33.34                                                             & -2.2                                                               & -17.22                                                             \\
\begin{tabular}[c]{@{}r@{}}Scenario 1a\\ vs.\\ Scenario1b\end{tabular} & -63.98                                                             & -5.57                                                              & -57.05                                                             & -51.95                                                             & 7.2                                                                & -52.97                                                            
\end{tabular}
\label{tab:4}
\end{table}

In Table \ref{tab:4}, we show the results for each branch in scenario 1a and 1b in comparison to the baseline method. Half of the trained models in scenario 1a show a performance increase compared to the baseline model. In scenario 1b, all models significantly outperform the baseline model and all models in scenario 1a except for branch 5. In branch 1 in scenario 1b we even observe a MAPE of 9.63\% that could be explained due to a “lucky shot” in that case. 

These results enable us to answer our first research question RQ 1 on how well we can forecast sales for single stores based on historic data. The results from scenario 1a, 1b and the baseline show that our method clearly outperforms the baseline method. In scenario 1b, we can significantly increase the performance of our method in comparison to the baseline method. This leads to the conclusion that more data leads to better performance, which both aligns with common sense and with previous research \cite{Banko2001}.

\section{Sales Forecasting with Transfer Machine Learning}

As we are able to create forecasting models that outperform simple baseline models and we see that models perform better when trained with more data, we use that knowledge as a basis for transfer learning. In the previous chapter, we conducted scenario 1a and 1b that answer RQ 1. However, our main interest in this work is the effectiveness and efficiency of transfer learning in the case of sales forecasting, which supports our theorizing process in regard to the generalization of forecast models. In this section, we first explain our method for different levels of adaption of transferred models and afterwards show and explain the obtained results of transfer machine learning without and with target adaption.

\subsection{Method}

We consider scenario 2 and 3 that aim towards showing the effectiveness and efficiency as depicted in Table 3. To show the effectiveness of transfer learning in our case, we have to demonstrate its feasibility. Thus, in scenario 2, we simulate a case, where no training data is available for the target branch, but a limited amount of data is available from a different one. To still make a prediction about the sales for the branch without data, a model is trained on data of a different branch and then used to make a forecast. The premise is that any data is better than no data (zero shot). Hereby, we apply transfer learning without any adaption of the target branch itself. In our case, a model for each branch is trained on the data until 2016 and each one of these models is then tested on the data from 2017 for any other branch. To answer RQ 2, we then compare the results of scenario 2 to the baseline method and the results in scenario 1a and 1b. Since we do not have any knowledge about the target branch scenario 2, we expect the results to be not as good as those of scenario 1a and 1b where we specifically trained models on a target’s historic data. However, we aim towards building models that achieve results comparable to baselines and models in other scenarios.  In that regard, we are able to evaluate the effectiveness of the transfer learning approach without any adaption in a case, where no training data is available for a target branch.

Furthermore, we assume that the prediction performance of a transfer learning approach can improve if transferred models are adapted to the target domain. Thus, in scenario 3, we consider that there is only little data available for the target domain (only data from 2016), but a longer history of sales data from other domains (until 2015). To realize that, we train models for each branch on data of a source branch until 2015, then, we use data of a target branch from 2016 to adapt the transferred model.  Finally, the adapted models are evaluated on the target data from 2017. We also compare the obtained results to the baseline and the other scenarios. In the following, we first show the results of scenario 2 (transfer learning without target adaption) and scenario 3 (transfer learning with target adaption).

\subsection{Results: Transfer Machine Learning without Target Adaption (Zero Shot)}

In Table \ref{tab:5} we show a matrix of results, where each $value_{i,j}$ represents the MAPE of a model that is trained on the branch  i (column) and tested on a branch j (row). Cases where i=j are ignored. Additionally, as we aim to find the best performing model for a branch k, we selected min($value_{k,j}$) for each branch and mark them in bold.

It is noticeable, that the results vary significantly. Whereas a model with branch 1 as a source and branch 6 as a target performs poorly (1005.63), a model that is originated in branch 5 reaches a MAPE of 37.09. We can observe that for branches of chain $\beta$ as a target, the average MAPE is higher—ranging from 205.00\% to 892.88\% — compared to chain $\alpha$, where the highest average MAPE of all transferred models per chain is 74.28\% (target branch 3). Additionally, also the standard deviation seems to be significantly higher for branches of chain $\beta$ as a target. For branch 5, we can observe a standard deviation of 758.09. Furthermore, branch 5 scores the highest MAPE with a model originating from branch 1 (chain $\alpha$). It is noticeable, that for all branches of chain $\alpha$, transferred models originating from the other chain $\beta$ are still performing well and even reach a MAPE at the minimum of 65.92\% (branch 4 to branch 3). However, as already stated out, we see the opposite effects with models, originating from branches of the first chain $\alpha$ that are tested on branches of the second chain $\beta$.

\begin{table}[htbp]
\setlength{\tabcolsep}{3pt}
\centering
\scriptsize
\caption{Results and their average and standard deviation of MAPE (in \%) for our method in scenario 2 (the lower, the better); best cases in bold }
\begin{tabular}{r|llllll|ll}
\textbf{\begin{tabular}[c]{@{}r@{}}Source\\ Target\end{tabular}} & \textbf{\begin{tabular}[c]{@{}l@{}}Ch. $\alpha$\\ Br. 1\end{tabular}} & \textbf{\begin{tabular}[c]{@{}l@{}}Ch. $\alpha$\\ Br. 2\end{tabular}} & \textbf{\begin{tabular}[c]{@{}l@{}}Ch. $\alpha$\\ Br. 3\end{tabular}} & \textbf{\begin{tabular}[c]{@{}l@{}}Ch. $\beta$\\ Br. 4\end{tabular}} & \textbf{\begin{tabular}[c]{@{}l@{}}Ch. $\beta$\\ Br. 5\end{tabular}} & \textbf{\begin{tabular}[c]{@{}l@{}}Ch. $\beta$\\ Br. 6\end{tabular}} & \textbf{AVG} & \textbf{SD} \\ \hline
\textbf{\begin{tabular}[c]{@{}r@{}}Ch. $\alpha$\\ Br. 1\end{tabular}}   & -                                                              & \textbf{40.98}                                                 & 49.66                                                          & 84.30                                                          & 94.25                                                          & 90.05                                                          & 71.84        & 24.66       \\
\textbf{\begin{tabular}[c]{@{}r@{}}Ch. $\alpha$\\ Br. 2\end{tabular}}   & 101.92                                                         & -                                                              & \textbf{16.78}                                                 & 67.34                                                          & 87.97                                                          & 79.12                                                          & 70.62        & 32.64       \\
\textbf{\begin{tabular}[c]{@{}r@{}}Ch. $\alpha$\\ Br. 3\end{tabular}}   & 108.92                                                         & \textbf{31.15}                                                 & -                                                              & 65.92                                                          & 87.38                                                          & 78.04                                                          & 74.28        & 28.79       \\
\textbf{\begin{tabular}[c]{@{}r@{}}Ch. $\beta$\\ Br. 4\end{tabular}}   & 465.57                                                         & 246.24                                                         & 202.68                                                         & -                                                              & 64.25                                                          & \textbf{43.29}                                                 & 205.00       & 169.35      \\
\textbf{\begin{tabular}[c]{@{}r@{}}Ch. $\beta$\\ Br. 5\end{tabular}}   & 1966.36                                                        & 1164.75                                                        & 999.62                                                         & 225.52                                                         & -                                                              & \textbf{108.15}                                                & 892.88       & 758.09      \\
\textbf{\begin{tabular}[c]{@{}r@{}}Ch. $\beta$\\ Br. 6\end{tabular}}   & 1005.63                                                        & 576.01                                                         & 493.71                                                         & 76.86                                                          & \textbf{37.09}                                                 & -                                                              & 437.86       & 398.59      \\ \hline
\textbf{AVG}                                                     & 729.68                                                         & 411.82                                                         & 352.49                                                         & 103.99                                                         & 74.78                                                          & 79.73                                                          &              &             \\
\textbf{SD}                                                      & 783.37                                                         & 475.31                                                         & 407.91                                                         & 68.34                                                          & 23.39                                                          & 23.69                                                          &              &            
\end{tabular}
\label{tab:5}
\end{table}

In Table \ref{tab:6} we show the results of the best performing models for every branch and compare them to the baseline method and the performances in scenario 1a and 1b. Although it is not possible to predetermine the best model in advance, the results shows the potential of the presented approach. As the models of scenario 2 gained no “knowledge” of the target branch and, hence, are not adjusted to them, most obtained results are not surprising. However, compared to scenario 1a and 1b, only some cases are similar in terms of performance. Although we observe, that in four out of six cases, models in scenario 2 perform worse than the baseline, compared to models in scenario 1a, zero shot models are in 50\% of all cases superior and even outperform the model of branch 2 in scenario 1b. Summarized, we reach respectable performances in terms of MAPE although the respective models are never trained on a target branch’s data.

\begin{table}[htbp]
\centering
\scriptsize
\setlength{\tabcolsep}{3pt}
\caption{Results of MAPE (in \%) for our method in scenario 2 compared to baseline and other scenarios. Comparisons in percentage points.}
\begin{tabular}{r|llllll}
\textbf{Method}                                                                & \textbf{\begin{tabular}[c]{@{}l@{}}Ch. $\alpha$\\ Br. 1\end{tabular}}        & \textbf{\begin{tabular}[c]{@{}l@{}}Ch. $\alpha$\\ Br. 2\end{tabular}}        & \textbf{\begin{tabular}[c]{@{}l@{}}Ch. $\alpha$\\ Br. 3\end{tabular}}        & \textbf{\begin{tabular}[c]{@{}l@{}}Ch. $\beta$\\ Br. 4\end{tabular}}        & \textbf{\begin{tabular}[c]{@{}l@{}}Ch. $\beta$\\ Br. 5\end{tabular}}         & \textbf{\begin{tabular}[c]{@{}l@{}}Ch. $\beta$\\ Br. 6\end{tabular}}        \\ \hline
\textbf{\begin{tabular}[c]{@{}r@{}}Scenario 2\\ best case\\ (BC)\end{tabular}} & \begin{tabular}[c]{@{}l@{}}40.98\\ (b2-\textgreater{}b1)\end{tabular} & \begin{tabular}[c]{@{}l@{}}16.78\\ (b3-\textgreater{}b2)\end{tabular} & \begin{tabular}[c]{@{}l@{}}31.15\\ (b2-\textgreater{}b3)\end{tabular} & \begin{tabular}[c]{@{}l@{}}43.29\\ (b6-\textgreater{}b4)\end{tabular} & \begin{tabular}[c]{@{}l@{}}108.15\\ (b6-\textgreater{}b5)\end{tabular} & \begin{tabular}[c]{@{}l@{}}37.09\\ (b5-\textgreater{}b6)\end{tabular} \\
\begin{tabular}[c]{@{}r@{}}Sc. 2 BC\\ vs. baseline\end{tabular}                & 92.66                                                                 & -37.10                                                                & -25.92                                                                & 61.59                                                                 & 235.86                                                                 & 24.54                                                                 \\
\begin{tabular}[c]{@{}r@{}}Sc. 2 BC\\ vs. Sc. 1a\end{tabular}                  & 53.25                                                                 & -35.88                                                                & -7.04                                                                 & 18.24                                                                 & 268.48                                                                 & -29.24                                                                \\
\begin{tabular}[c]{@{}r@{}}Sc. 2 BC\\ vs. Sc. 1b\end{tabular}                  & 325.54                                                                & -32.09                                                                & 116.46                                                                & 146.10                                                                & 243.44                                                                 & 50.46                                                                
\end{tabular}
\label{tab:6}
\end{table}

The performance of a zero shot scenario is similar if we consider predictions six months into the future: In comparison to the baseline method, the best models in scenario 2 perform better in 50\% of the cases, in case of branch 2 even outperforms the baseline (-54.73\%), scenario 1a (-31.87\%) and 1b (-24.75\%) and in case of branch 6, the best model outperforms scenario 1a by -4.8\%.

We conduct scenario 2 to answer, whether or not it is possible to transfer models from one source branch to a different one as a target, without having any historic data of the target branch available (RQ 2). As we reach respectable performances with models in scenario 2 and outperform one third of the baseline models, 50\% of the models in scenario 1a, and even one model of scenario 1b, we conclude, that a zero shot prediction based on transfer learning can be feasible.

\subsection{Results: Transfer Learning with Target Adaption}

As we already obtained good results in scenario 2, we expect an increase of performance in scenario 3, as in this case, models are trained on a source branch until 2015, adapted to data of 2016 and finally tested on the targets data of 2017. 

\begin{figure}[htbp]
    \centering
    Scenario 1b
	\includegraphics[width=0.85\linewidth]{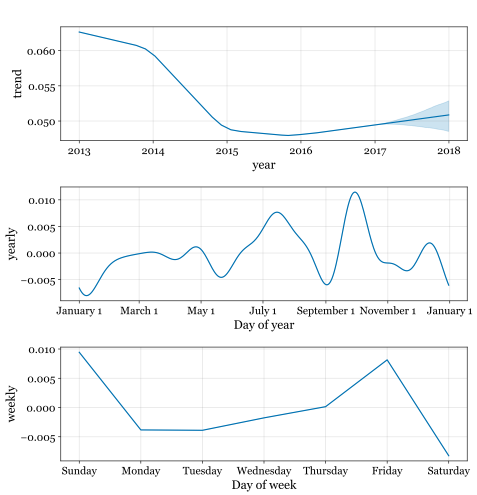}
	Scenario 3
	\includegraphics[width=0.85\linewidth]{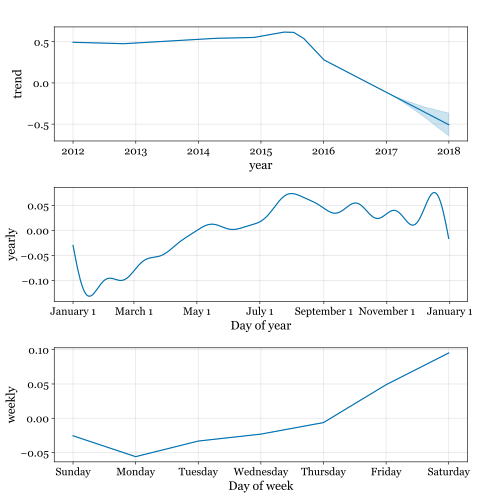}
	\caption{Components of the model for branch 4 in scenario 1b (top) vs. adapted model (source: branch 1, target: branch 4) in scenario 3 (bottom)}
	\label{fig:4}       
\end{figure}

In Figure \ref{fig:4}  we depict the components of the additive models for target branch 4 (top) in scenario 1b and scenario 3 (bottom) to see the change of trend and weekly as well as monthly seasonality curves. Note, that in this case the adapted model is originally trained on data of chain $\alpha$ and then adapted to a branch of chain $\beta$, where seasonality and trend are differing (see Figure \ref{fig:1}). In the trend component of scenario 3, we observe a decline in 2016 due to the adaption on the target set. By comparing the yearly seasonality, there is a clear change resembling the seasonality in the time series of branch 4 as depicted in Figure \ref{fig:1}. Looking at the monthly seasonality in scenario 3, we see similarities to patterns in the seasonality of branch 1 and branch 4. Additionally, there is a change of weekly seasonality towards increased sales on Saturdays, which also supports the hypothesis that the transfer learning adaption works as intended.

\begin{figure}[htbp]
    \centering
	\includegraphics[width=0.7\linewidth]{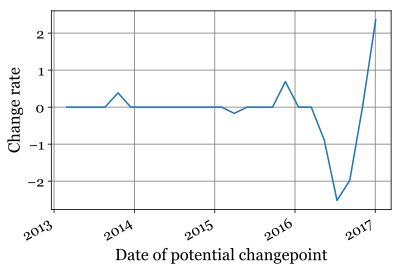}
	\caption{Changepoint weights of model trained on data of branch 1 until 2015, adapted on data from branch 4 from 2016}
	\label{fig:5}       
\end{figure}

These plots for the trained model imply that with only one year of adaption on the target branch, the model could have gained enough insight on the target branch’s seasonality. The adapted model on one year of data achieved to display a similar yearly seasonality plot like a model specifically build on four years of a targets historic data, which has four years more of data. In Figure \ref{fig:5}, the weights of learned changepoints for the model originated in branch 1 and adapted on branch 4 in scenario 3. The figure shows that changepoints in the period of adaption have a higher weight than others. Thus, we can deduce that our model learned that the data structure has changed and an adaption is necessary by increasing the changepoint weights.

In contrast to scenario 2, results in scenario 3 are more stable in terms of standard deviation (see Table \ref{tab:7}). Whereas in the previous situation the standard deviation of target branches reaches a high of 758.09, now across all branches standard deviations are more consistent between 13.46 (branch 1) and 23.49 (branch 6) Considering the average performances per target branch, we not only see a strong performance increase in comparison to scenario 2, but also a narrow window of results between 31.16\% (branch 1) and 46.32\% (branch 3).  

Adapted models that origin in a branch of the same chain as the target branch, perform better than others. The results show that with exception of branch 4, best performing adapted models for a branch are always originating from another branch in the same chain. By comparing the average and standard deviation of a column, we can get a clue on the suitability of a branch as an origin for transferring a model. In that regard, we observe that branch 2 seems to reach stable and good results with an average of 23.37\% and a low standard deviation of 3.15. In contrast, branch 1 reaches an average of 51.10\% and a standard deviation of 21.20, which still represents good results in comparison to previous evaluations in other scenarios. 

\begin{table}[htbp]
\setlength{\tabcolsep}{3pt}
\centering
\scriptsize
\caption{Results of MAPE (in \%) for our method in scenario 3}
\begin{tabular}{r|llllll|ll}
\textbf{\begin{tabular}[c]{@{}r@{}}Source\\ Target\end{tabular}} & \textbf{\begin{tabular}[c]{@{}l@{}}Ch. $\alpha$\\ Br. 1\end{tabular}} & \textbf{\begin{tabular}[c]{@{}l@{}}Ch. $\alpha$\\ Br. 2\end{tabular}} & \textbf{\begin{tabular}[c]{@{}l@{}}Ch. $\alpha$\\ Br. 3\end{tabular}} & \textbf{\begin{tabular}[c]{@{}l@{}}Ch. $\beta$\\ Br. 4\end{tabular}} & \textbf{\begin{tabular}[c]{@{}l@{}}Ch. $\beta$\\ Br. 5\end{tabular}} & \textbf{\begin{tabular}[c]{@{}l@{}}Ch. $\beta$\\ Br. 6\end{tabular}} & \textbf{AVG} & \textbf{SD} \\ \hline
\textbf{\begin{tabular}[c]{@{}r@{}}Ch. $\alpha$\\ Br. 1\end{tabular}}   & -                                                              & 19.18                                                          & 17.98                                                          & 28.26                                                          & 46.89                                                          & 43.49                                                          & 31.16        & 13.46       \\
\textbf{\begin{tabular}[c]{@{}r@{}}Ch. $\alpha$\\ Br. 2\end{tabular}}   & 24.96                                                          & -                                                              & 13.61                                                          & 33.63                                                          & 53.25                                                          & 61.15                                                          & 37.32        & 19.68       \\
\textbf{\begin{tabular}[c]{@{}r@{}}Ch. $\alpha$\\ Br. 3\end{tabular}}   & 31.43                                                          & 21.69                                                          & -                                                              & 40.14                                                          & 68.93                                                          & 69.43                                                          & 46.32        & 21.86       \\
\textbf{\begin{tabular}[c]{@{}r@{}}Ch. $\beta$\\ Br. 4\end{tabular}}   & 62.09                                                          & 23.71                                                          & 57.80                                                          & -                                                              & 38.67                                                          & 39.80                                                          & 44.41        & 15.60       \\
\textbf{\begin{tabular}[c]{@{}r@{}}Ch. $\beta$\\ Br. 5\end{tabular}}   & 69.22                                                          & 27.55                                                          & 63.35                                                          & 23.25                                                          & -                                                              & 36.58                                                          & 43.99        & 21.01       \\
\textbf{\begin{tabular}[c]{@{}r@{}}Ch. $\beta$\\ Br. 6\end{tabular}}   & 67.80                                                          & 24.73                                                          & 71.34                                                          & 21.66                                                          & 38.76                                                          & -                                                              & 44.85        & 23.49       \\ \hline
\textbf{AVG}                                                     & 51.10                                                          & 23.37                                                          & 44.81                                                          & 29.38                                                          & 49.30                                                          & 50.09                                                          &              &             \\
\textbf{SD}                                                      & 21.20                                                          & 3.15                                                           & 26.97                                                          & 7.62                                                           & 12.55                                                          & 14.38                                                          &              &            
\end{tabular}
\label{tab:7}
\end{table}

Similar to scenario 2, we select the best performing adapted models to compare them to the baseline and results in other scenarios. The baseline models get outperformed by the best models in scenario 3. In scenario 1a we apply the same method but train only on data of 2016. In comparison, we pre-train models in scenario 3 on other branches, transfer and adapt those on the target data of 2016. As shown in Table \ref{tab:8}, scenario 3 models clearly outperform the baseline for every branch. Similarly, all models in scenario 1a are surpassed, in case of branch 6 even by a 58.67\% decrease of MAPE. Thus, the results indicate, that in a case where little data is available for a target branch—in case of the baseline and scenario 1a data from 2017—it can be significantly advantageous to use data originating form similar branches to enhance model performance using transfer learning. 

By comparing scenario 3 to scenario 1b, we observe in three cases a decrease of performance up to 86.70\%.  This could have many reasons as, for instance, that a transfer in those cases might not be suitable. The target branches and all available source branches could be too different in terms of structure or underlying patterns. However, in three cases model performances in scenario 3 by far surpass those in scenario 1b, and in case of branch 2 the performance reaches a peak of 13.61—representing a 44.92\% MAPE decrease.

When compared to scenario 2, we see that all models are showing an increased performance, as there is no adaption to the target branch in scenario 2. 

\begin{table}[htbp]
\centering
\scriptsize
\caption{Results of MAPE (in \%) for our method in scenario 3 compared to baseline and other scenarios. Comparisons in percentage points.}
\begin{tabular}{r|llllll}
\textbf{Method}                                                                & \textbf{\begin{tabular}[c]{@{}l@{}}Ch. $\alpha$\\ Br. 1\end{tabular}} & \textbf{\begin{tabular}[c]{@{}l@{}}Ch. $\alpha$\\ Br. 2\end{tabular}} & \textbf{\begin{tabular}[c]{@{}l@{}}Ch. $\alpha$\\ Br. 3\end{tabular}} & \textbf{\begin{tabular}[c]{@{}l@{}}Ch. $\beta$\\ Br. 4\end{tabular}} & \textbf{\begin{tabular}[c]{@{}l@{}}Ch. $\beta$\\ Br. 5\end{tabular}} & \textbf{\begin{tabular}[c]{@{}l@{}}Ch. $\beta$\\ Br. 6\end{tabular}} \\ \hline
\textbf{\begin{tabular}[c]{@{}r@{}}Scenario 3\\ best case\\ (BC)\end{tabular}} & \textbf{17.98}                                                          & \textbf{13.61}                                                          & \textbf{21.69}                                                          & \textbf{23.71}                                                          & \textbf{23.25}                                                          & \textbf{21.66}                                                          \\
\begin{tabular}[c]{@{}r@{}}Sc. 3 BC\\ vs. baseline\end{tabular}                & -15.46                                                         & -48.98                                                         & -48.98                                                         & -11.49                                                         & -27.79                                                         & -27.26                                                         \\
\begin{tabular}[c]{@{}r@{}}Sc vs.\\ 3 BC, S1a\end{tabular}                     & -32.75                                                         & -35.37                                                         & -35.27                                                         & -35.23                                                         & -20.78                                                         & -58.67                                                         \\
\begin{tabular}[c]{@{}r@{}}Sc vs.\\ 3 BC, S1b\end{tabular}                     & 86.70                                                          & -44.92                                                         & 50.72                                                          & 34.79                                                          & -26.16                                                         & -12.12                                                         \\
\begin{tabular}[c]{@{}r@{}}Sc vs.\\ 3 BC, S2\end{tabular}                      & -56.12                                                         & -18.89                                                         & -30.36                                                         & -45.22                                                         & -78.50                                                         & -41.60                                                        
\end{tabular}
\label{tab:8}
\end{table}

We conducted scenario 3 to answer the question of how well we can forecast sales with adapted transferred models (RQ 3). Our results indicate, that a transfer and adaption of models across branches of different restaurant chains can enhance forecast performance depending on the source and target branch. We see a significant performance increase in comparison to a baseline method and in a scenario with a limited amount of data. Complying with our previous insights, transfer learning is suitable in cases where no or little data is available. Furthermore, we observe that some branches are on average performing better and more stable as a source for transferred models in terms of performance than other branches.

\section{Discussion and Conclusion}


In this work, we analyze the generalization of analytical models through transfer learning for sales forecasting for regression based algorithms. We first evaluate the effectiveness and efficiency for a sales forecasting approach based on the data of single branches for two separate restaurant chains and are able to outperform every regarded baseline. Then, we evaluate the effect of increasing the training data and find, that more training data leads to better results. 

On the basis of these results with “isolated” machine learning, we realize the transfer of forecasting models across branches and chains—and evaluate the effectiveness and efficiency of transfer learning with two levels of adaption. 
First, we analyze a scenario, where no data is available for a target branch and, thus, we build a forecasting model first on data of a source branch, and then test its performance on a target branch without any adaption. As we do this for every combinations of branches, we obtain multiple results in our evaluation. We show that already transferred models without adaption to a target branch deliver good results and, in some cases, they even outperform the baseline models that are dedicatedly built on a target branch’s data. Furthermore, we observe, that some branches serve better as a source than others, and for some branches, transferred models perform significantly worse. It is noticeable, that even across restaurant chains, transferred models are able to reach a fairly solid performance even though the restaurant chains serve different types of food. Thus, we show the effectiveness of transfer learning in our case. 

To further adapt the transferred models and, thus, generalize analytical models, we adapt transferred models on a subset of the target’s data and repeat this procedure for every combination of two branches in our set. The obtained results clearly show, that adapted transferred models not only outperform the baseline method, but also most other models in scenarios that we considered in our study. Furthermore, some adapted transferred models outperform models, that are trained in an isolated manner for a branch.


Therefore, we contribute to the body of knowledge by showing the suitability of transfer learning in general and extending its capabilities by using adaptive learning to reach superior performances. Furthermore, we show the applicability of transfer learning for shallow learning techniques. In most cases our transfer learning with adaption is able to outperform any baseline---even those which had all historical data of a single entity available. These insights help us to establish important foundations in previously uncharted territory. The application of transferable models to the challenge of time series in general and sales forecasting in particular. 

These results have several significant implications for research and practice. First of all, as the results from the case are very promising, transfer learning may be a valuable addition to the tool set of data scientists. We are able to show that multiple models are able to capture generalizable knowledge across different cases and entities. These insights fit well into the Artificial Intelligence (AI) research priorities of Russell et al. (2015) \cite{Russell2015}, stressing the importance of research dealing with the economic impact of machine learning and AI. If we are able to successfully capture abstract knowledge in machine learning models across different problems and entities, we can imagine digital “markets” for the exchange of models, which require no or little additional training. 

Such a transfer has, furthermore, the advantage that only abstract models are exchanged. Depending on the used algorithm, reconstruction of an original training set is impossible (e.g., ANN) and, therefore, an exchange of models is feasible from a data governance point of view. This insight specially addresses the lack of data exchange due to its confidentiality \cite{Mitchell2006}.



Besides these contributions, this work has several limitations. With the current state of the research, we only consider a transfer between two branches and, thus, do not utilize the full potential of all data that is available across branches in the entire “system”. There could be options on transferring models across various branches and by that, further generalize and enhance sales forecasting models. 

Additionally, we only evaluate one type of adaption of transferred models to a target domain but there are various techniques of adaption that depend on certain base techniques of forecasting. We do not vary training, adaption and testing phase. However, by dynamically combining those phases across branches, it might be possible to further enhance model performance. 
Furthermore, it still needs more work on understanding why there are differences in performance depending on a source and target branch. We select the best performing transferred models after testing every option between a source and target branch. Thus, the question arises, how to determine beforehand which branches are suitable for a transfer and which are not. As a starting point, characteristics of branches (e.g., location, number of tables) \cite{Karb2020} or analytical models \cite{hirt2020sequential} could be considered as indicators for transferability.

In terms of our sales forecasting method, we are solely considering sales data to make a forecast. By considering different external input source, such as weather or nearby events, forecasting performance could be further improved. 

As we only take a first step towards generalization of models in sales forecasting by considering two branches at a time, studies should focus on how to generalize models across an arbitrary number of branches. That would enable to move towards a more dynamic exchange of machine learning models. To untap the potential of transferring machine learning models across various application domains, studies are needed that address dynamic and automated adaption of transferred models.  

\bibliographystyle{IEEEtran}
\bibliography{References.bib}

\end{document}